# An Open Natural Language Processing (NLP) Framework for EHR-based Clinical Research: A Case Demonstration Using the National COVID Cohort Collaborative (N3C)

Sijia Liu[1,*], Andrew Wen[1,*], Liwei Wang[1,*], Huan He[1,*], Sunyang Fu[1,*], Robert Miller[2], Andrew Williams[2], Daniel Harris[3], Ramakanth Kavuluru[3], Mei Liu[4], Noor Abu-el-rub[4], Dalton Schutte[5], Rui Zhang[5], Masoud Rouhizadeh[6], John D. Osborne[7], Yongqun He[8], Umit Topaloglu[9], Stephanie S Hong[10], Joel H Saltz[11], Thomas Schaffter[12], Emily Pfaff[13], Christopher G. Chute[10], Tim Duong[14], Melissa A. Haendel[15], Rafael Fuentes[16], Peter Szolovits[17], Hua Xu[18], Hongfang Liu[1], National COVID Cohort Collaborative (N3C) Natural Language Processing (NLP) Subgroup, National COVID Cohort Collaborative (N3C)

1. Department of Artificial Intelligence and Informatics, Mayo Clinic.
2. Tufts Clinical and Translational Science Institute, Tufts Medical Center.
3. Department of Internal Medicine, University of Kentucky.
4. Department of Internal Medicine, University of Kansas Medical Center.
5. Department of Pharmaceutical Care & Health Systems, University of Minnesota at Twin Cities.
6. Department of Pharmaceutical Outcomes & Policy, University of Florida
7. Department of Computer Science, University of Alabama at Birmingham.
8. University of Michigan Medical School.
9. Wake Forest Baptist Medical Center.
10. Department of Medicine, Johns Hopkins University.
11. Department of Biomedical Informatics, Stony Brook University.
12. Sage Bionetwork.

13. Department of Medicine, University of North Carolina Chapel Hill.
14. Albert Einstein College of Medicine.
15. Center for Health AI, University of Colorado Anschutz Medical Campus.
16. Alex Informatics.
17. Department of Electrical Engineering and Computer Science, Massachusetts Institute of Technology.
18. School of Biomedical Informatics, University of Texas Health Science Center at Houston.

* These authors contributed equally to this study

# Abstract

Despite recent methodology advancements in clinical natural language processing (NLP), adoption of clinical NLP models within the clinical and translational research community remains hindered by issues with ETL process heterogeneity and human factor variations. In this study, we proposed an open NLP development framework with the aim of addressing these issues. The viability of such a platform was evaluated on a COVID-19 use case through sites participating in the National COVID Cohort Collaborative (N3C). As part of our assessment of the impact of single vs. multi-site NLP algorithm development, we evaluated the performance of both an NLP ruleset developed solely using a single site's clinical narratives as well as one further refined using a synthetic derived dataset sourced from three sites (Mayo, UKen, and UMN). The single-site ruleset resulted in performances of 0.876, 0.706, and 0.694 in F-scores for Mayo, Minnesota, and Kentucky test datasets, respectively, while the multi-site NLP ruleset improved performances to 0.884, 0.769 and 0.806. The results of our use case test run inform us of the importance of a multi-site federated development, evaluation, and implementation framework. As such, we aim to meet this need with our framework by providing the tools necessary to conduct NLP development in a collaborative manner through consensus building, process coordination, and best practice sharing.

## Introduction

Over the past decade, Electronic Health Record (EHR) systems have been increasingly implemented at US healthcare institutions. Large amounts of detailed longitudinal patient information, including lab tests, medications, disease status, and treatment outcomes, have consequently been accumulated and made electronically available. These large clinical databases are valuable data sources for clinical and translational research. As a result, major initiatives have been established to exploit this crucial resource, including the Clinical and Translational Science Awards (CTSA) Program's National Center for Data to Health (CD2H)/National COVID Cohort Collaborative (N3C)[1,2], the Electronic Medical Records and Genomics (eMERGE) Network[3], the Patient-Centered Outcomes Research Institute's (PCORI) Clinical Research Networks (CRNs)[4], the NIH All of Us Research Program[5], and the Observational Health Data Science and Informatics (OHDSI) Consortia with demonstrated successes[6,7,8,9].

One common challenge faced by those initiatives is, however, the prevalence of clinical information embedded in unstructured text[10]. Compared to structured data entry, text is a more conventional way in the healthcare environment to document impressions, clinical findings, assessments, and care plans. Even with the advent of sophisticated EHR systems, studies have shown that capturing health information fully in structured format through data entry is unlikely to happen and a blended model where physicians use templates when and where possible and dictate the details of a patient visit in text[11].

Natural language processing (NLP) has been promoted as having a great potential to extract information from text[12]. NLP algorithms can generally be categorized into using either symbolic or statistical methods[13]. Since the turn of the century, machine learning algorithms (i.e., statistical NLP) have gained increased prominence in clinical NLP research[14]. Nevertheless, a substantial portion of clinical NLP use cases leverages symbolic techniques given that dictionary or rule-based methodologies suffice to meet the information needs of many clinical applications under specific use cases. In the context of EHR-based clinical research, NLP has been leveraged to assist information extraction and knowledge conversion at different stages of research including feasibility assessment, eligibility criteria screening, data elements extraction, and text data analytics. As a result, an increasing number of clinical research benefits from state-of-the-art NLP solutions and have been reported ranging from disease study areas[15, 16, 17, 18] to drug-related studies[19, 20]. A majority of existing clinical NLP studies are, however, done within a mono-institutional environment[13], which may suffer from limited external validity and research inclusiveness. Compared with single-site research, multisite research potentially offers larger sample size, more adequate representation of participant demographics (e.g., age, gender, race, ethnicity, and social-economic status), and more diverse investigator expertise, which may ultimately yield a higher level of research evidence[21, 22, 23, 24].

Despite a plethora of recent advances in adopting NLP for clinical research, there have been barriers towards adoption of NLP solutions in clinical and translation research, especially in multi-site settings. The root causes of these barriers can be categorized into two major reasons: 1) heterogeneity of ETL (extract, transform, load) processes between differing sites with their own disparate EHR environments, and 2) human factor variation in gold standard corpus development processes.

**ETL Process Heterogeneity.** The challenges faced by NLP development and evaluation to facilitate the secondary use of EHR data originate from the complex, voluminous, and dynamic nature of the data being documented and stored within a heterogeneous set of disparate, institution specific, EHR implementations. Variations in EHR system vendors, data infrastructure (e.g., unified, ontology driven, and de-centralized), and institutions' modes of operation can lead to idiosyncratic ways of clinical documentation, transformed, and representation[25]. Collecting these data would require a significant expenditure of effort to locate, retrieve, and link EHR data into a specific format[26]. This variability in ETL processes required to support a high level of data heterogeneity brings additional challenges in the adoption of NLP for clinical and translational research, which substantially limits both the cross-institutional interoperability of developed NLP solutions and the reproducibility of the associated evaluations.

**Human factor variation in gold standard corpus development process.** The process of developing, evaluating, and deploying NLP solutions in both mono- and multi-site environments can be task-specific, iterative, and complex, often involving a multitude of stakeholders with diverse backgrounds[13, 26]. A key step prior to model development is corpus annotation, the process of developing a gold standard by marking the occurrence of both task-defined sets of clinical information as well as their associated interpretative linguistic features (e.g., certainty, status) within text documents. Due to the complexity of clinical language, creating such gold standard corpora requires significant expenditure of domain expertise and time as clinical experts regularly make decisions directly affecting study cohort, annotation guideline, and task definitions. Studies have discovered potential biases in clinical decision making and interpretation of clinical guidelines[27], in coding of clinical terminologies[28], and in interpretation of imaging findings[29]. This issue can be further exacerbated when conducting multi-site collaborations due to inter-site variations in care practice[30, 31], ultimately affecting the validity and reliability of the resulting gold standard corpus. A coordinated, transparent, and collaborative platform is therefore needed to promote open team science collaboration in NLP algorithm development and evaluation through consensus building, process coordination, and best practice sharing.

Built upon our previous work[32, 33], here, we proposed an open NLP development framework to address the aforementioned issues through the following components: 1) an interoperable NLP infrastructure for incorporation of different NLP engines utilizing a clinical common data model for data source interfacing and representation with the aim of reducing the impact of the heterogeneity of ETL processes; 2) a transparent multi-site participation workflow on corpus development and evaluation with the aim of addressing the variation in data abstraction and annotation processes between sites; and 3) a user-centric crowdsourcing interface for collaborative ruleset development that enables effectively and efficiently gathering, synthesizing, and fusing site-specific knowledge and findings. To demonstrate the viability of the framework, we conducted a case study where we developed, evaluated, and implemented an NLP algorithm for extracting [34, 35, 36] COVID-19 signs and symptoms to support the National COVID Cohort Collaborative (N3C).

# Results

## Framework Description

The framework itself consists of a data ingestion layer, a processing layer, and a data persistence layer. The architecture of the proposed framework is illustrated in Figure 3. The **data ingestion layer** works as the data collector with the ability to read text from a configurable variety of data sources such as relational databases or file systems including and load them into the NOTE table of OMOP CDM. The **processing layer** serves as the NLP engine where information extraction from raw texts happens given a set of heuristic rules created by various NLP engines. By default, as an example implementation, the MedTagger[37] NLP engine is provided, although alternative NLP engines can be substituted by wrapping their respective NLP pipelines to conform to a provided API specification. After the term modifiers added by contextual rules from ConText Algorithm[38] around the extracted condition mentions, these conditions will compose clinical events with temporal information. The reason we opt for a symbolic solution is due to its simplicity, transparency, and interpretability as the outcomes are fully deterministic based on the definition of the rules. When the baseline rulesets and dictionaries are made available to the public, they can therefore be easily refined by different users from different sites. The **data persistence**

**layer** stores resulting extracted NLP artifacts in the OMOP CDM NOTE_NLP table as the events are extracted from NLP systems.

The framework is distributed as open-source software under the Apache 2.0 license via Github in three parts: 1) ETL Backbone (https://github.com/OHNLP/Backbone) with an example NLP engine (https://github.com/OHNLP/MedTagger), 2) process documentation (https://github.com/OHNLP/N3C-NLP-Documentation/wiki), 3) open-source collaborative platform for developing NLP rulesets (https://github.com/OHNLP/OHNLPTK). The demo homepage (Figure 2(a) - https://ohnlp4covid-dev.n3c.ncats.io/) demonstrates the N3C NLP engine outputs on annotating clinical text using the baseline rulesets and dictionary. The annotations are from components of Sign/Symptom extractor, temporal information extractor and dictionary lookup extractor. To further customize each model, the users can visit "Rule Editor" (https://ohnlp4covid-dev.n3c.ncats.io/ie_editor) and the "Dictionary Builder" (https://ohnlp4covid-dev.n3c.ncats.io/dict_builder) page (Figure 2(b)). Figure 2(c) provides an example of the rules editing interface with the baseline COVID-19 ruleset. The rulesets can be tested in real time by clicking the "Upload and test" button, where the rulesets will be uploaded, and the NLP engine will be generated for testing and debugging purposes. As a use case study, we also provide an example NLP project for extracting signs/symptoms related to COVID-19 that was developed as an example use case for this framework. The elements with original texts such as text snippets and concept mentions are truncated before submission.

## N3C Case Study

**NLP Algorithm Development and Evaluation**: Table 1 shows the annotation corpora statistics. A COVID-19 sign/symptom ruleset was produced consisting of 17 concepts. The IAA of the annotated corpus was 0.686 F1-score for Mayo, 0.516 for UMN and 0.211 for UKen. Two NLP algorithms were evaluated in this study. One was developed based solely on the narratives sourced from a single site (Mayo Clinic). The other used the resulting NLP algorithm from the single site and fine-tuned based on the annotated training data from an additional two sites (UMN and UKen). Table 2 shows the performance of the single-site NLP algorithm and Table 3 shows the performance of the multi-site NLP algorithm. The single-site ruleset resulted in performances of 0.876, 0.706, and 0.694 in F-scores for Mayo, Minnesota, and Kentucky test datasets, respectively, while the multi-site NLP ruleset improved performances to 0.884, 0.769 and 0.806. The performance of the multi-site NLP algorithm was better than that of the single-site NLP algorithm, but both showed a degrading trend from Mayo site to other sites.

Tables 4, 5 and 6 show the results of error analysis for the three sites. For FP, major discrepancies between the NLP algorithm and the gold standard were due to the NLP algorithm extracting mentions that are not COVID signs/symptoms but for instruction/patient education, adverse events/indication of treatment, clinical goal/precaution, template, etc. It should be noted that gold standards were not always correct, and in some notes, it was hard to judge if the mentions are COVID signs/symptoms when symptoms are not appearing with COVID or de-identified dates are inconsistent. For FN, reasons include NLP algorithm not complete, tokenization error due to de-identification process, template, and annotation errors.

# Discussion

In this study, we proposed an open NLP development framework with the following properties: an interoperable NLP infrastructure, a transparent multi-site participation workflow, and a user-centric crowdsourcing interface. The key goal of this framework is to facilitate multi-site collaborative development, evaluation, and implementation of NLP algorithms. The framework has been implemented to support efforts conducted by the National COVID Cohort Collaborative (N3C) to enable the utilization of unstructured text in high throughput.

Here, we have presented our results from running our framework using a centralized annotation process on texts sourced from multiple sites after de-identification, with the aim of assessing the impact on NLP algorithm development (single-site algorithm vs multi-site algorithm). Several pragmatic implementation challenges were discovered that may impact the intermediate and final NLP results. We observed that IAA varied greatly between the three sites despite the fact that annotators had been trained using de-identified Mayo notes (0.686 F1-score for Mayo, 0.516 for UMN, 0.211 for UKen). Firstly, utilizing a centralized annotation approach, the process of text data collection took a very long time because each site needs to complete de-identification before sharing data. Secondly, it was a challenge for annotators to work on annotation tasks that spanned a long period of time. Thirdly, the shared data sets were usually small, and as such, annotators had no chance to do annotation training using these outside notes, and it was hard for them to get familiar with the disparate variety of document structures from other sites.

Both multi-site and single-site NLP algorithms showed a degrading trend in performance from Mayo site to other sites, albeit this issue being less prominent in the multi-site NLP algorithm as compared to the single-site NLP algorithm. The data sharing issues also impacted NLP algorithm performance. First, training sets from outside institutions were very small due to the small number of shared notes, causing difficulties in developing comprehensive rules as features, patterns, and contextual information that could appear in third party narratives could not be fully represented in such a small sample. Second, de-identification processes could cause text span issues that may impact the input text format and thus NLP algorithm performance. The algorithm performance for algorithms developed through a centralized mode was therefore not ideal for immediate use at multiple sites, as additional local fine-tuning is still needed before final implementation and application.

Our experiment results showed that a centralized approach towards multi-site NLP algorithm development is suboptimal for advancing the adoption of NLP techniques in the clinical and translational research community, this further support our proposed federated method. The experiment also demonstrates that deployment of NLP algorithms for multi-site studies needs to be done in each local site. To ensure the scientific rigor of the data generated, each site need to perform annotation and evaluation on their own while collectively contributing to NLP algorithm development and refinement. Since the NLP models are to be shared in rule-based systems, the models can be shared without the concerns typically associated with language resources involving the Protected Health Information (PHI) issue.

In the proposed workflow, each site will evaluate the NLP algorithms for concept extraction by creating a gold standard corpus based on the common annotation guidelines. The federated evaluation can be deployed leveraging cloud computing through a centralized controller where NLP algorithms can be distributed to each institution. NLP Sandbox[1] is an example of such an evaluation framework, which uses Docker[39] containers to encapsulate algorithm implementations. By adopting this process, the evaluation only happens behind each institution's firewall, and only the summary statistics on NLP algorithm performance (i.e., no raw data containing PHI) is transferred out of the firewall. Performance statistics, such as the precision, recall, and F1-score, as defined depending on the experimental setting, can be obtained in near real-time and can thus be used as part of continuous development workflows.

This federated process offers several benefits. For instance, when conducting error analysis, we discovered that contexts played an important role in this case study. Error analyses showed it was not a trivial task to extract COVID signs/symptoms, as their occurrence is not necessarily isolated only to occurring due to COVID, and as they could appear as adverse events/indication of treatment, or in instruction/patient education, or clinical goal/precaution, etc. This posed a challenge not only for annotation, but also for the NLP algorithm development. One benefit of the federated annotation and development process is that these contexts can be systematically incorporated by local expertise in the annotation process.

---

[1] NLP Sandbox: https://github.com/nlpsandbox

Deployment of a federated development framework requires the participation of multiple sites. Adoption can, however, be hindered by the fact that the process of translating NLP algorithms into implementation is complex, much like the "bench to bedside" process that translates laboratory discoveries into patient care. To facilitate participation in our federated method, we have developed a further suite of tools such as MedTator[40] and best practice guidelines[41]. MedTator, a serverless annotation tool, aims to provide an intuitive and interactive user interface for high-quality annotation corpus generation. The best practice guideline contains detailed instructions for facilitating multisite annotation practice with the following key activities: task formulation, cohort screening, annotation guideline development, annotation training, annotation production, and adjudication.

Simply having the toolsets be available, is, however, insufficient. Pragmatically, we have seen that there is a hyper focus on novel methods in academia with competing as opposed to collaborative priorities in NLP algorithm development. Our experience suggests that a collaborative development process for NLP algorithms is needed for truly implementable and useful multi-site NLP solutions. This is one of the key goals we seek to achieve with the Open Health Natural Language Processing (OHNLP) Collaboratory and have thus positioned our framework's workflow to facilitate this task. Additionally, we recognize that it is not simply a software problem, a local workforce is also needed at each institution. As a consequence of conducting coordinated development of NLP algorithms deployed using our framework as a solution for consortia-specific tasks such as with the N3C, we simultaneously build the human workforce locally at institutions necessary to conduct the federated development, evaluation, and implementation of NLP algorithms using our framework.

# Methods

## Design Principles

**Incorporating standards and interoperability.** A common barrier to the widespread adoption of NLP in clinical research is the need to transform input and outputs to conform to part of an overall pipeline. While seemingly straightforward, such a task is difficult without prior significant investment in associated infrastructure and dedicated software development. It is therefore desirable to leverage existing infrastructure where possible and incorporate such an effort into the distributed NLP pipeline to reduce technical burden on the end user.

There is, however, significant variation in terms of available infrastructure and data availability amongst different institutions. Creating a solution that is immediately suitable for all these environments out of the box would be immensely challenging. For that reason, we sought to leverage existing data modeling efforts that are likely to be already adopted by academic medical institutions to standardize the data ingestion and output process. In our implementation, we chose the Observational Health Data Sciences and Informatics' Observational Medical Outcomes Partnership common data model (OHDSI/OMOP CDM) to handle input of clinical narratives via the NOTE table and output via the NOTE_NLP table. This brings the advantage that input/output is now standardized: so long as institutions have already transformed their clinical data into the OMOP CDM, and/or their downstream NLP-reliant applications read from the OMOP CDM database, no additional technical development burden is needed.

It is important to note that standardization as a default only serves to simplify adoption for those who already have a solution complying with the standard and cannot be a comprehensive solution. A purely OMOP CDM reliant solution is not ideal, as not all institutions will have their own OMOP CDM instance and standing up such an instance to just use a pipeline may produce undue burden. For that reason, input/output in our infrastructure is modularized, and can be substituted at will: the default OMOP CDM I/O utilizes a variant of SQL-based data extractors/writers, and the specific query and connection strings used can be substituted via plaintext configuration changes. Additionally, SQL-based I/O is not the only supported setting, a variety of other data sources including Elasticsearch, google cloud storage, amazon s3, and plaintext are included as well as configuration-swappable options.

**Crowdsourcing algorithm development.** To promote collaboration and sharing efforts between participants in the algorithm development process, we built a crowdsourcing platform for domain experts to upload, customize, and examine their NLP algorithms in an interactive web application. Users can create keyword-based and rule-based algorithms and test the performance in the online environment instantly. The crowdsourcing platform consists of three modules based on our NLP system to support expert collaboration, including dictionary builder, regular expression rule set editor, and detection result visualization.

The dictionary builder can extend the keyword collection used by the algorithm. Users can customize particular terms from the ontology database such as CIDO [42] and MONDO [43]. The regular expression rule set editor provides an integrated interface to help users customize their own regular expression rule set (on top of an existing dictionary, if desired), to support use cases such as extraction of new symptoms, treatments, or outcomes. The detection result visualization is designed based on Brat annotation tool [44] to check the results generated by different methods.

## Case study

The National COVID Cohort Collaborative[36] (N3C) is a novel partnership that includes the Clinical and Translational Science Awards (CTSA) Program hubs, the National Center for Advancing Translational Science (NCATS), the Center for Data to Health (CD2H) and the community, focusing on collaborative sharing of structured EHR data. Access to unstructured data is limited due to protection of PHI and clinical care decision logics, that were further contributing to NLP infrastructure lacking within the consortia. However, structured data does not show the whole picture from the EHR perspective, greatly restricting research activities. In this case study, extraction of COVID-19 signs and symptoms was used as a case study to investigate the viability of the proposed framework among sites participating in the N3C.

**Centralizing gold standard corpus development.** Due to resources and time constraints at each of the N3C sites, we opted to conduct the gold standard corpus development process in a centralized manner. A collection of de-identified and synthesized clinical documents was gathered from participating sites through an existing de-identification effort led by the NCATS Clinical Data to Health (CD2H). The N3C deidentification and synthetic text generation workflow is illustrated in Figure 1. Specifically, clinical notes from patients with positive COVID-19 test results from three institutions, Mayo Clinic, the University of Kentucky (UKen), and the University of Minnesota at Twin Cities (UMN) were initially collected. Notes that were not office visit notes (e.g., nurse calls, etc.), notes that had fewer than 1000 characters, and notes that were authored more than 14 days prior to the date of the patient's earliest positive COVID-19 test result were further filtered out. A total of 369 clinical notes from these sites that met these criteria were randomly selected, de-identified using the de-identification program developed by the Medical College of Wisconsin followed by manual review. The removed PHI identifiers are replaced by the programmatically added synthetic texts. We collected 20 signs and symptoms of COVID-19 as a basic COVID-19 concept set according to the recommendations from the CDC and Mayo Clinic. Five out of the 20 concepts are emergency warning signs including dyspnea, chest pain, delirium, hypersomnia and cyanosis. We then gathered formal definitions of each clinical concept from the Coronavirus Infectious Disease Ontology (CIDO) [42]. Based on the Open Biological and Biomedical Ontology (OBO) Foundry library, CIDO concepts were imported from 45 ontologies, and it uses Human Phenotype Ontology (HPO) [45] for phenotypes. Some representative phenotypes shown in COVID-19 have been imported to the CIDO. However, if the chosen COVID-19 clinical concepts were not collected by the CIDO, we re-pulled them from the HPO to the CIDO. We also gathered cross-reference concept codes from the CIDO including UMLS [46], SNOMED-CT [47], MeSH [48], HPO, MeDDRA [49].

We selected available clinical notes from both inpatients and outpatients in the two-week window preceding the order date of the first positive COVID-19 result as the annotation cohort. After the text data was collected from participating sites, the same annotation process was completed by the annotator team from Mayo Clinic to generate the gold standard annotations on COVID-19 signs and symptoms. There are 313 clinical notes from Mayo Clinic, 20 notes from UKen and 36 notes from UMN. Annotators were first trained using Mayo notes to gain better understanding of the annotation guidelines. Inter-annotator agreement (IAA) was calculated after annotation and corresponding discrepancies were resolved by discussions between the two annotators to generate a final gold standard dataset.

**NLP algorithm development and evaluation.** Using the annotated corpus, we developed both a single-site and multi-site NLP algorithm using a regular expression-based matching method, which has been widely adopted for information extraction in clinical settings. Specifically, for the Mayo data, we randomly chose 101 notes out of the 313 annotated notes as development set, 105 notes as validation set, and the remaining 107 notes were used as the testing set. For the UKen data, 10 notes were used for training and 10 for testing. For the UMN data, 18 was used for training and 18 for testing. Single-site algorithm was developed using the development set and validation set from Mayo, tested on the Mayo testing set and all data from UKen and UMN. Multi-site algorithm was generated through further refinement of the single-site algorithm using training sets from UKen and UMN and then tested on testing sets from all sites.

We evaluated the performance of single-site and multi-site algorithms using precision, recall, and F1-score for the annotated concept mentions, without and with certainty. A span can be represented from the start position to the end position of the concept mention. Certainty is an attribute of the concept mention including positive, negated, hypothetical and possible. For the mention-level evaluation without certainty, when there are overlaps between the gold standard mention span and the NLP detected mention span while the concept type (i.e., the specific sign/symptom such as fever, cough) is the same, it is considered a true positive (TP). If a concept mention exists in the gold standard annotation but not detected by the NLP algorithm, or spans overlap but the concept type is not matched, it is considered as a false negative (FN). If a concept mention is detected by the algorithm but does not exist in the gold standard annotation, the concept is considered as a false positive (FP). For the mention-level span and certainty evaluation, certainty match needs to be considered when calculating TP, FN and FP. The precision, recall and F1-score are then calculated as follows. We further manually analyzed errors from multi-site algorithm mention-level evaluation without certainty.

$$\Pr ecision = \frac{TP}{TP + FN}$$

$$Recall = \frac{TP}{TP + FN}$$

$$F1 = \frac{2 \cdot \Pr ecision \cdot Recall}{\Pr ecision + Recall}$$

# Acknowledgment

This research was possible because of the patients whose information is included within the data and the organizations and scientists who have contributed to the on-going development of this community resource https://doi.org/10.1093/jamia/ocaa196. The analyses described in this publication were conducted with data or tools accessed through the NCATS N3C Data Enclave https://covid.cd2h.org and N3C Attribution & Publication Policy v1.2-2020-08-25b and supported by NCATS U24 TR002306. This task was made possible by the National Center for Advancing Translational Sciences of the National Institutes of Health under award number U01TR02062 and the Bill & Melinda Gates Foundation. The content is solely the responsibility of the authors and does not necessarily represent the official views of the National Institutes of Health.

# Code Availability Statement

Framework components can be found on GitHub:

- ETL Pipeline: https://github.com/OHNLP/Backbone
- NLP Implementation: https://github.com/OHNLP/MedTagger
- Web Rule Editor Front-end: https://github.com/OHNLP/OHNLPTK
- MedTator annotation tool: https://github.com/OHNLP/MedTator

The developed NLP ruleset can be found at
https://github.com/OHNLP/covid19ruleset/tree/main/covid19

# Data Availability Statement

- A detailed annotation guideline outlining the goals of the NLP task and how the corpora were annotated can be found at https://github.com/OHNLP/N3C-NLP-Documentation/wiki/Annotation-guideline-for-COVID-19-concepts
- The sample deidentified synthetic corpus used as part of this study can be found at https://github.com/OHNLP/N3C-NLP-Documentation/blob/master/n3c_omop_sample.csv

# Competing Interests Declaration

MAH has a founding interest in Pryzm Health. HX and The University of Texas Health Science Center at Houston have financial related interests at Melax Technologies Inc.

# Author contributions

Project Conceptualization: SL, AWen, LW, HH, SF, HL

Data curation: SL, AWen, LW, HH, RM, AWilliams, DH, RK, ML, NA, MR, RZ, JDO, JHS

Data integration: AWen, LW, HH, RM, DH, RK, ML, NA, RZ, TS, YH, EP, SSH, CGC, JHS

Data analysis: SL, AWen, LW, RM, RK, NA, RZ, MR, TS

Software development: SL, AWen, HH, MR, TS

Data quality assurance: SL, RM, DH, RK, LM, NA, RZ, TS, JDO, HY, EP, TD, PS, HX

Draft the manuscript: SL, AWen, LW, HH, RM, RZ, HL, SF

Critical revision of the manuscript for important intellectual content: AWilliams, RK, ML, NA, YH

Project evaluation: LW, SF, RM, AWilliams, DH, RK, ML, NA, RZ, MR, TS

Project management: SL, LW, RZ, TS, EP, JHS, RF, HL

Regulatory oversight / admin: EP, CGC, HL

Database / Information systems admin: RM, DH, NA, RZ, TD

Biological subject matter expertise: LW, YH, UT, MAH

Funding acquisition: MAH, CGC, PS, HX, HL

# Figures

Figure 1. N3C deidentification and synthetic text generation workflow

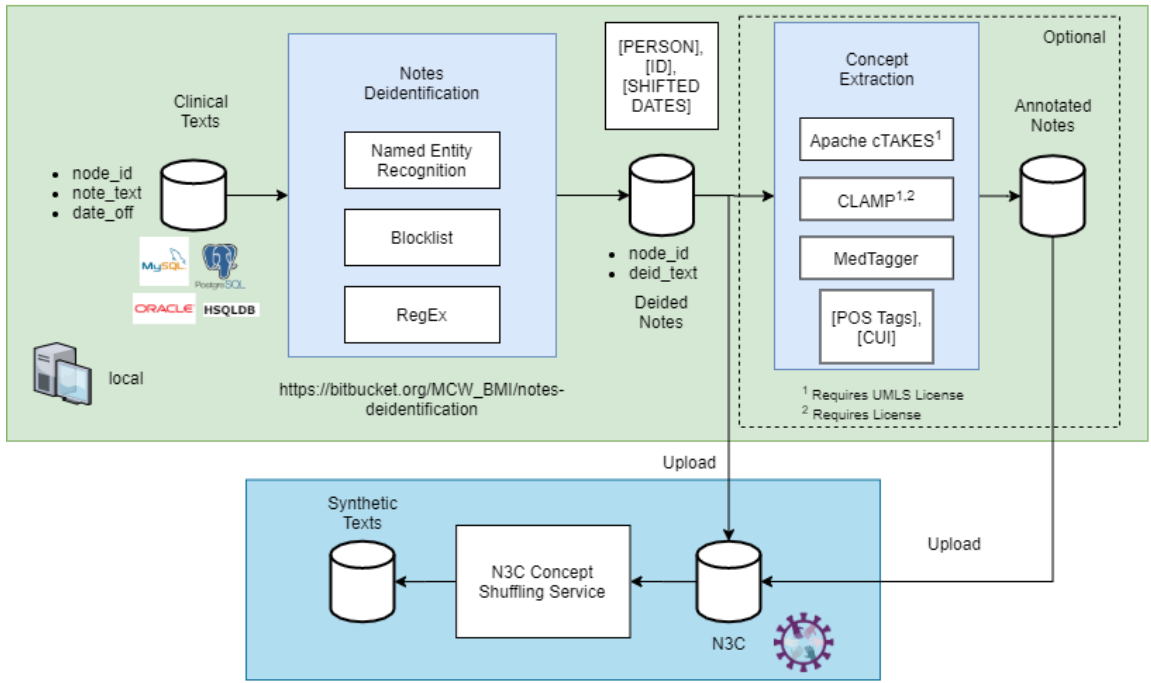

Figure 2. Screenshots of the Web GUI

(a)

(b)

Screenshot of the N3C NLP Engine Demo Rule Editor interface, showing the Rule Package tab with UI Mode options (Easy Mode, Full Mode), Ruleset options (Local, New, COVID-19), Ruleset Name field, Download (As ZIP), Upload (Test Rules), and Help (About) buttons. The Easy Rule Editor panel displays Contexts with a contextRule containing regex patterns, and a Rule List (19 rules) with checkboxes for DYSPNEA, NASAL_OBSTRUCTION, LOSS_OF_APPETITE, ELEVATED_LDH, ELEVATED_PT_TIME, LYMPHOPENIA, DIARRHEA, ABDOMINAL_PAIN, SORE_THROAT, HEADACHE, MYALGIA, FATIGUE, DRY_COUGH, FEVER, INFLUENZA_EXPL, PATCHY_INFILTRATES, GROUND_GLASS_INFILTRATES, REMOVE_VACCINE, REMOVE_CAREPLAN. Concept Mention boxes show synonyms for DYSPNEA (dyspnea, shortness of breath, sob, respiratory difficulty, breathing difficulties, difficult breathing, difficult to breathe, panting, hyperpnea, gasping, gasp, rapid breathing), NASAL_OBSTRUCTION (blocked nose, congested nose, nose block, nasal congestion, stuffy nose, runny nose, nasal blockage, blockage of nose, obstruction of nose, nasal obstruction, congestion of nose), LOSS_OF_APPETITE (loss of appetite, appetite loss, no appetite, anorexia, food aversion), ELEVATED_LDH (elevated lactate dehydrogenase, elevated LDH, LDH increased, lactate dehydrogenase increased, lactate dehydrogenase activity increased), ELEVATED_PT_TIME (elevated PT, prothrombin increased, prolonged pt, prolonged prothrombin time, abnormal prothrombin time, abnormal pt), LYMPHOPENIA (lymphopenia, lymphocytopenia, decreased lymphocyte, decreased lymphocytes, low lymphocyte count, low lymphocyte number).

(c)

Screenshot of N3C NLP Engine Demo — Dictionary Builder interface showing an Ontology Tree (CIDO, 6818 classes) on the left with checkboxes for entities such as continuant, occurrent, process, host response to coronavirus infection, disorder prevention, infectious disease epidemic, vaccine-induced host response, exposure event or process, Physiological Effects [PE], life cycle, host exposure to infectious agent, coronaviral process to host, life-death temporal boundary, Cellular or Molecular Interactions [MoA], immunization against infectious agent, immune response, host-coronavirus interaction, immunization, life cycle stage, Clinical Kinetics [PK], disease course; and a Term List on the right with definitions for continuant, occurrent, process, host response to coronavirus infection, disorder prevention, and infectious disease epidemic.

Figure 3. Diagram of N3C NLP Solution

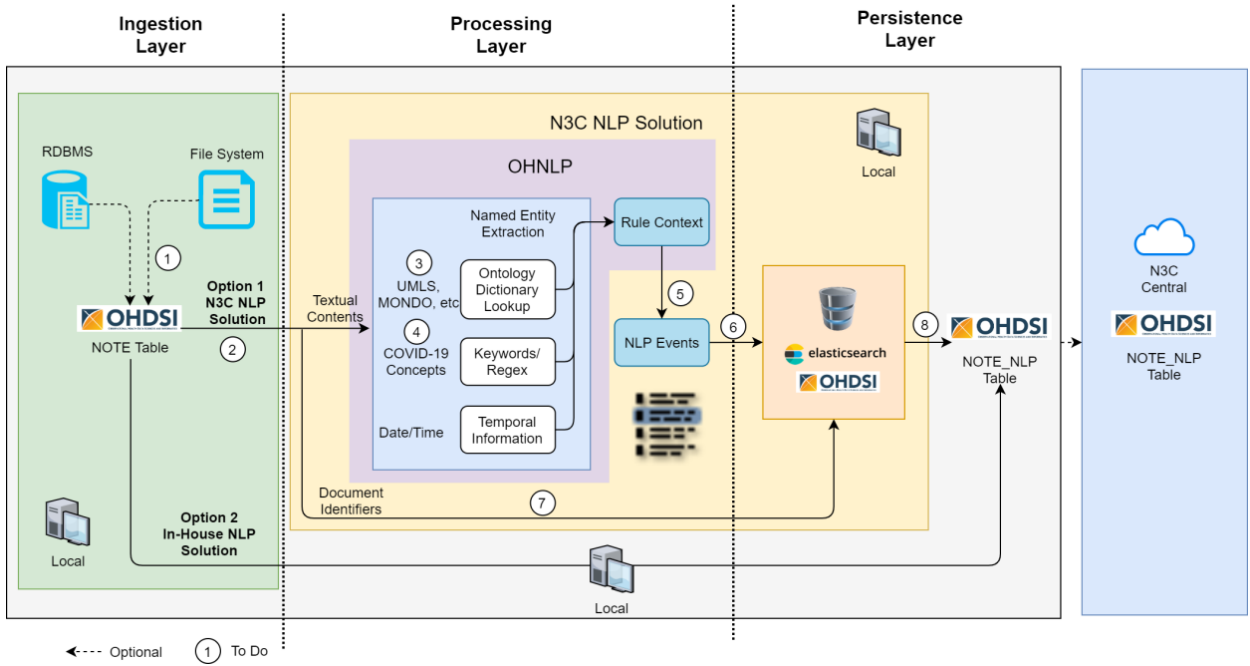

# Tables

Table 1. Annotation corpora statistics. (Mayo: Mayo Clinic, UKen: University of Kentucky, UMN: University of Minnesota)

| Concepts | Mayo (313 notes) | UKen (20 notes) | UMN (36 notes) |
|---|---|---|---|
| Abdominal_pain | 59 | 2 | 3 |
| Chest_pain | 62 | 2 | 11 |
| Chill | 51 | 6 | 6 |
| Cough | 104 | 14 | 43 |
| Cyanosis | 9 | 4 | 17 |
| Delirium | 38 | 2 | 10 |
| Diarrhea | 92 | 5 | 11 |
| Dyspnea | 199 | 19 | 46 |
| Fatigue | 61 | 15 | 13 |
| Fever | 148 | 25 | 53 |
| Headache | 43 | 6 | 15 |
| Hypersomnia | 6 | 0 | 14 |
| Loss_of_appetite | 41 | 2 | 4 |
| Loss_of_smell | 23 | 4 | 6 |
| Loss_of_taste | 19 | 2 | 5 |
| Myalgia | 21 | 6 | 8 |
| Nasal_obstruction | 16 | 6 | 14 |
| Nausea | 87 | 7 | 10 |
| Sore_throat | 16 | 4 | 17 |
| Vomiting | 86 | 6 | 14 |

Table 2. Performance of the single-site NLP algorithm (Mayo: Mayo Clinic, UKen: University of Kentucky, UMN: University of Minnesota)

| Dataset | Span | | | Span+Certainty | | |
|---|---|---|---|---|---|---|
| | Precision | Recall | F1 | Precision | Recall | F1 |
| Mayo | 0.882 | 0.869 | 0.876 | 0.789 | 0.639 | 0.706 |
| UKen | 0.698 | 0.714 | 0.706 | 0.664 | 0.643 | 0.653 |
| UMN | 0.658 | 0.735 | 0.694 | 0.534 | 0.438 | 0.481 |

Table 3. Performance of the multi-site NLP algorithm (Mayo: Mayo Clinic, UKentucky: University of Kentucky, UMN: University of Minnesota)

| Dataset | Span | | | Span+Certainty | | |
|---|---|---|---|---|---|---|
| | Precision | Recall | F1 | Precision | Recall | F1 |
| Mayo | 0.863 | 0.908 | 0.884 | 0.824 | 0.681 | 0.746 |
| UKen | 0.696 | 0.859 | 0.769 | 0.662 | 0.734 | 0.696 |
| UMN | 0.718 | 0.918 | 0.806 | 0.562 | 0.456 | 0.504 |

Table 4. Error analysis of the multi-site algorithm mention-level evaluation without certainty for Mayo site

| Error types of FP | No. FP (%) | Error types of FN | No. FN (%) |
|---|---|---|---|
| Annotation error: missing annotation | 17 (26%) | NLP algorithm not complete | 21 (66%) |
| Hard to judge if are COVID signs/symptoms | 15 (23%) | Annotation error | 8 (25%) |
| Not COVID signs/symptoms - instruction/patient education | 10 (15%) | Tokenization error due to input format | 2 (6%) |
| Not COVID signs/symptoms - adverse events/indication of treatment | 7 (11%) | Template | 1 (3%) |
| Not COVID signs/symptoms - others (anesthesia plan, symptoms of other disease, etc.) | 5 (7%) | | |
| Not COVID signs/symptoms - clinical goal/precaution | 4 (6%) | | |
| Not COVID signs/symptoms - template | 5 (7%) | | |
| NLP algorithm not precise | 2 (3%) | | |

Table 5. Error analysis of the multi-site algorithm mention-level evaluation without certainty for UMN

| Error types of FP | No. FP (%) | Error types of FN | No. FN (%) |
|---|---|---|---|
| Not COVID signs/symptoms - instruction/patient education | 30 (61%) | NLP algorithm not complete | 11 (85%) |
| Not COVID signs/symptoms - (substance use history; overdose; due to surgery, other comorbidity, etc.) | 7 (14%) | Annotation error | 2 (15%) |
| Annotation error: missing annotation | 6 (12%) | | |
| Not COVID signs/symptoms - template | 4 (8%) | | |
| Hard to judge if are COVID signs/symptoms | 2 (3%) | | |

Table 6. Error analysis of the multi-site algorithm mention-level evaluation without certainty for UKen

| Error types of PF | No. FP (%) | Error types of FN | No. FN (%) |
|---|---|---|---|
| Not COVID signs/symptoms - (substance use history; overdose; due to surgery, other comorbidity, etc.) | 8 (33%) | NLP algorithm not complete | 7 (78%) |
| Annotation error: missing annotation | 5 (21%) | Annotation error | 2 (22%) |
| Not COVID signs/symptoms - instruction/patient education | 5 (21%) | | |
| Hard to judge if are COVID signs/symptoms | 3 (13%) | | |
| Not COVID signs/symptoms - template | 2 (8%) | | |
| NLP algorithm not precise | 1 (4%) | | |